\documentclass[11pt,a4paper]{article}
\usepackage[hyperref]{acl2017}
\usepackage{times}
\usepackage{latexsym}
\usepackage{graphicx}
\usepackage{url}

\usepackage{multirow}
\usepackage{hhline}

\usepackage{graphicx}
\usepackage{epstopdf}
\usepackage{subfig}
\usepackage{mdwlist}
\usepackage{enumitem}
\usepackage{amsfonts}
\usepackage{amsmath}
\usepackage{mathtools}
\usepackage{setspace}
\usepackage{stackengine}
\usepackage{array}
\usepackage{comment}

\aclfinalcopy % Uncomment this line for the final submission
\def\aclpaperid{***} %  Enter the acl Paper ID here

\title{Understanding and Detecting Supporting Arguments of Diverse Types}
\author{Xinyu Hua \and Lu Wang \\ College of Computer and Information Science \\ Northeastern University \\ Boston, MA 02115 \\ {\tt hua.x@husky.neu.edu} \quad {\tt luwang@ccs.neu.edu}}

\begin{document}
\maketitle

\begin{abstract}
\fontsize{10}{12}\selectfont
%Constructing arguments of high quality requires gathering diverse types of information, ranging from facts to opinions. However, retrieving effective supporting information is a challenging process. 
We investigate the problem of sentence-level \textit{supporting argument detection} from relevant documents for user-specified claims. A dataset containing claims and associated citation articles is collected from online debate website \url{idebate.org}. We then manually label sentence-level supporting arguments from the documents along with their types as \textsc{study}, \textsc{factual}, \textsc{opinion}, or \textsc{reasoning}. 
We further characterize arguments of different types, and explore whether leveraging type information can facilitate the supporting arguments detection task. 
Experimental results show that LambdaMART~\cite{burges2010ranknet} ranker that uses features informed by argument types yields better performance than the same ranker trained without type information.
  
  %We study the problem of sentence-level supporting argument detection for user-specified claims in debatabase\footnote{www.idebate.org}, by leveraging sentence type information based on their subjectivity and credibility. As a pilot study we annotate a medium-sized dataset with type information. And we propose a feature-rich ranking framework to retrive supporting argument. Experiments show that salient features exhibit significantly different distribution among different types with real-world implications, and by incorporating the type information our ranker achieves higher performance.
\end{abstract}

\section{Introduction}
% if we want to understand how human construct arguments, as a first step, we need to know what kind of information they seek for to support their claim and how to automatically detect the supporting arguments

%Constructing arguments of high quality would require (1) extracting diverse supporting information from relevant documents, and (2) organizing it into a coherent structure with proper substantiation and reasoning. 

%study: 108, 9.76% factual: 575, 51.94% opinion: 382, 34.51% reasoning: 42, 3.79%

Argumentation plays a crucial role in persuasion and decision-making processes. An argument usually consists of a central claim (or conclusion) and several supporting premises. Constructing arguments of high quality would require the inclusion of diverse information, such as factual evidence and solid reasoning~\cite{rieke1997argumentation,park2014identifying}. 
For instance, as shown in Figure~\ref{fig:intro}, the editor on \url{idebate.org} -- a Wikipedia-style website for gathering pro and con arguments on controversial issues, utilizes arguments based on study, factual evidence, and expert opinion to support the anti-gun claim ``legally owned guns are frequently stolen and used by criminals". 
However, it would require substantial human effort to collect information from diverse resources to support argument construction. In order to facilitate this process, there is a pressing need for tools that can automatically detect supporting arguments. 

%\XY{For instance, the editor on \url{idebate.org} -- a Wikipedia-style website for gathering pro and con arguments on controversial issues, utilizes arguments based on study, factual evidence, and expert opinion to support the anti-gun claim ``legally owned guns are frequently stolen and used by criminals":}

%\vspace{-2mm}
\begin{figure}[h]
%\captionsetup{font=small}
	\fontsize{10}{12}\selectfont
    \setlength{\tabcolsep}{0.8mm}
    
%    \hspace{3mm}
	\begin{tabular}{p{74mm}}

        - A June 2013 IOM report states that ``almost all guns used in criminal acts enter circulation via initial legal transaction''. [study] \\
        - Between 2005 and 2010, 1.4 million guns were stolen from US homes during property crimes (including bulglary and car theft), a yearly average of 232,400. [factual] \\
        - Ian Ayres, JD, PhD, \ldots states, ``with guns being a product that can be easily carried away and quickly sold at a relatively high fraction of the initial cost, the presence of more guns can actually serve as a stimulus to burglary and theft.'' [expert opinion]\\

	\end{tabular}
	\caption{\fontsize{10}{12}\selectfont Three different types of arguments used to support the claim ``Legally owned guns are frequently stolen and used by criminals". }
\label{fig:intro}
\end{figure}

%\XY{As we can tell, it would require substantial human effort to collect information from diverse resources to support argument construction. In order to facilitate this process, there is a pressing need for tools that can automatically detect supporting arguments.} 
%
To date, most of the argument mining research focuses on recognizing argumentative components and their structures from constructed arguments based on curated corpus~\cite{mochales2011argumentation,stab2014identifying,feng2011classifying,habernal-gurevych:2015:EMNLP,nguyen-litman:2016:P16-1}. Limited work has been done for retrieving supporting arguments from external resources. Initial effort by~\newcite{rinott2015show} investigates the detection of relevant factual evidence from Wikipedia articles. However, it is unclear whether their method can perform well on documents of different genres (e.g. news articles vs. blogs) for detecting distinct types of supporting information.

\begin{figure}[t]
%\captionsetup{font=small}
	\fontsize{10}{12}\selectfont
	%\small
	\setlength{\tabcolsep}{0.8mm}
	\begin{tabular}{|p{75mm}|}
	\hline
	- \textbf{Topic}: This house would ban cosmetic surgery \\
	- \textbf{Claim}: An outright ban would be easier than the partial bans that have been enacted in some places. \\
	- \textbf{Human Constructed Argument}: \color{blue}{$\ldots$This potentially leaves difficulty drawing the line for what is allowed.[1]} $\ldots$ \\ 		\hline
	\end{tabular}

	%\vspace{-3mm}
	%\hspace{-2mm}
	\begin{tabular}{|p{75mm}|}
	\hline
	\underline{\textbf{Citation Article}}\\
	
	[1]: {\it ``Australian State Ban Cosmetic Surgery for Teens"}\\
	
	%\ldots\\
	- \ldots It is unfortunate that a parent would consider letting a 16-year-old daughter have a breast augmentation."\vspace{.1cm}
	
	{\color{blue}\textit{- But others worry that similar legislation, if it ever comes to pass in the United States, would draw a largely arbitrary line -- and could needlessly restrict some teens from procedures that would help their self-esteem.}}\vspace{.1cm}
	
	- Dr. Malcolm Z. Roth, director of plastic surgery at Maimondes Medical Center in Brooklyn, N.Y. , said he believes that some teens are intelligent and mature enough to comprehend the risks and benefits of cosmetic surgery.\ldots\\

	\hline
	\end{tabular}
	%\vspace{-3mm}
	\caption{\fontsize{10}{12}\selectfont
	A typical debate motion consists of a Topic, Claims, and Human Constructed Arguments. Citation article is marked at the end of sentence. Our goal is to find out supporting argument (in \textit{italics}) from citation article that can back up the given claim.}
	%based on the Human Constructed Argument citing this article. }
	%An example of annotation. Topic, claim, and argument are taken from \url{idebate.org}. Annotators first check the topic, claim and argument constructed by human editors, they then need to find the sentence(s) in cited article that editors used for argument}
\label{fig:example_intro}
\end{figure}

%In this work, we present a new corpus, which support our study to detect arguments of different types from 
In this work, \textit{we present a novel study on the task of sentence-level supporting argument detection from relevant documents for a user-specified claim}. Take Figure~\ref{fig:example_intro} as an example: assume we are given a claim on the topic of ``banning cosmetic surgery" and a relevant article (cited for argument construction), we aim to automatically pinpoint the sentence(s) (in \textit{italics}) among all sentences in the cited article that can be used to back up the claim. We define such tasks as \textit{supporting argument detection}. 
Furthermore, another goal of this work is to understand and characterize different types of supporting arguments. Indeed, human editors do use different types of information to promote persuasiveness as we will show in Section~\ref{sec:data}. Prediction performance also varies among different types of supporting arguments. 
%\XY{And we also observe there are intrinsic differences in terms of the difficulties to analyze and detect different types of arguments.}
%\XY{As we expect human editors employ different types of arguments as strategies to promote persuasiveness and coherence.} \LW[add more explanation?]

% data to collect, annotation, some statistics
%Given that none of the existing datasets is suitable for our study, we first collect a corpus from Idebate, which contains hundreds of contentious debate topics and corresponding claims from both pro and con sides. As is shown in Figure~\ref{fig:example_intro}, each claim is supported with a paragraph of human constructed argument, with cited articles marked on sentence level. We collect all topics and claims along with associated citation articles, and recruit annotators to manually identify the supporting arguments. 
Given that none of the existing datasets is suitable for our study, we collect and annotate a corpus from Idebate, which contains hundreds of debate topics and corresponding claims.\footnote{The labeled dataset along with the annotation guideline will be released at \url{xyhua.me}.} 
As is shown in Figure~\ref{fig:example_intro}, each claim is supported with some human constructed argument, with cited articles marked on sentence level. 
After careful inspection on the supporting arguments, we propose to label them as \textsc{study}, \textsc{factual}, \textsc{opinion}, or \textsc{reasoning}. Substantial inter-annotator agreement rate is achieved for both supporting argument labeling (with Cohen's $\kappa$ of 0.8) and argument type annotation, on 200 topics with 621 reference articles.
%\XY{We achieve satisfactory inter-annotator agreement score for both supporting argument labeling (0.80) and type annotation (0.75).
%We achieve an inter-annotator agreement of 0.80 on supporting argument labeling, and as high as 0.75 for type annotation, which are reasonably good considering the complexity of the task. Annotators also discuss afterwards to resolve the disagreement.\footnote{Our dataset and annotations will be made publicly available upon publication.}
%We collected labels for a subset of these revisions: label 200 topics and 621 reference articles, with 1107 labeled sentences with meticulous disagreement resolution. our inter rater agreement score shows steady increase after each round and finally ends up at 0.78, which is reasonably good considering the complexity of the task.}

% news articles is a major source, many study and reasoning from scientific, opinions and reasoning from blogs
% propose features to characterize different types of supporting argument
% predict on types of candidate sentences, and see if add type information will help
Based on the new corpus, we first carry out a study on characterizing arguments of different types via type prediction. We find that arguments of \textsc{study} and \textsc{factual} tend to use more concrete words, while arguments of \textsc{opinion} contain more named entities of person names. 
We then investigate whether argument type can be leveraged to assist supporting argument detection. Experimental results based on LambdaMART~\cite{burges2010ranknet} show that utilizing features composite with argument types achieves a Mean Reciprocal Rank (MRR) score of 57.65, which outperforms an unsupervised baseline and the same ranker trained without type information. Feature analysis also demonstrates that salient features have significantly different distribution over different argument types. 
%\XY{[which outperforms non-trivial baseline based on word embedding similarity (47.65) and the same ranker trained without type information (56.43).]}
%\XY{which outperforms an unsupervised baseline using word embedding similarity (47.65), and the same ranker trained without compositing type information. (56.43)}
%Feature analysis shows that \LW{add} \XY{salient features have significantly different distribution over different types of arguments. }
%
%This also points out several promising directions for future work. \LW{add info} 

For the rest of the paper, we summarize related work in Section~\ref{sec:related}. The data collection and annotation process is described in Section~\ref{sec:data}, which is followed by argument type study (Section~\ref{sec:type}). Experiment on supporting argument detection is presented in Section~\ref{sec:relevance}. We finally conclude in Section~\ref{sec:conclusion}.

\section{Related Work}
\label{sec:related}
% belong to the arising field of argument mining, but existing work focuses on detecting argument components and their structure from the same documents
Our work is in line with argumentation mining, which has recently attracted significant research interest. Existing work focuses on argument extraction from news articles, legal documents, or online comments without given user-specified claim~\cite{moens2007automatic,palau2009argumentation,mochales2011argumentation,park2014identifying}.  
%\XY{The earliest work on argument scheme dates back to Ancient Greece, as in \cite{aristotle2006rhetoric}, where persuasion is categorized into \textit{logos, ethos}, and \textit{pathos}, based on different functions they perform in rhetoric structures.} 
Argument scheme classification is also widely studied~\cite{biran2011identifying,feng2011classifying,rooney2012applying,stab2014identifying,alkhatib-EtAl:2016:COLING}, which emphasizes on distinguishing different types of arguments. To the best of our knowledge, none of them studies the interaction between types of arguments and their usage to support a user-specified claim. This is the gap we aim to fill.

\section{Data and Annotation}
\label{sec:data}
We rely on data from \url{idebate.org}, where human editors construct paragraphs of arguments, either supporting or opposing claims under controversial topics. We also extract textual citation articles as source of information used by editors during argument construction. 
In total we collected 383 unique debates, out of which 200 debates are randomly selected for study. After removing  invalid ones, our final dataset includes 450 claims and 621 citation articles with about 53,000 sentences. 

\noindent \textbf{Annotation Process.} 
As shown in Figure~\ref{fig:example_intro}, we first annotate which sentence(s) from a citation articles is used by the editor as supporting arguments. Then we annotate the type for each of them as \textsc{study}, \textsc{factual}, \textsc{opinion}, or \textsc{reasoning}, based on the scheme in Table~\ref{tab:annoscheme}.\footnote{We end up with the four-type scheme as a trade-off between complexity and its coverage of the arguments.} 
For instance, the highlighted supporting argument in Figure~\ref{fig:example_intro} is labeled as \textsc{reasoning.}

\begin{table}[t]
\fontsize{10}{12}\selectfont
\begin{tabular}{p{73mm}}
\hline
\textsc{Study:} Results and discoveries, usually quantitative, as a result of some research investment.\\% For example, results of some experiments or poll. \\ 
\textsc{Factual:} Description of some occurred events or facts, or chapters in law or declaration. \\%Usually can be obtained without research investment. For example, description of objective environment, issuance of law, historical events.\\
\textsc{Opinion:} Quotes from some person or group, either direct or indirect. It usually contains subjective, judgemental and evaluative languages, and might reflect the position or stance of some entity. \\ %For example, comments on laws and policies from officials, speculation or prediction on stock markets.\\
\textsc{Reasoning:} Logical structures. It usually can be further broken down into causal or conditional substructures. \\ %For example, to explain how oil extraction could break ecosystem by giving causal chains and their effects.\\  
\hline
\end{tabular}
\caption{\fontsize{10}{12}\selectfont Annotation scheme for our dataset. Due to space limit, we do not show detailed explanations and examples. }
\label{tab:annoscheme}
\vspace{-.4cm}
\end{table}

Two experienced annotators were hired to identify supporting arguments by reading through the whole cited article and locating the sentences that best match the reference human constructed argument. 
This task is rather complicated since human do not just repeat or directly quote the original sentences from citation articles, they also paraphrase, summarize, and generalize. 
For instance, the original sentence is ``The global counterfeit drug trade, a billion-dollar industry, is thriving in Africa", which is paraphrased to ``This is exploited by the billion dollar global counterfeit drug trade" in human constructed argument.

The annotators were asked to annotate independently, then discuss and resolve disagreements and give feedback about current scheme. 
We compute inter-annotator agreement based on Cohen's $\kappa$ for both supporting arguments labeling and argument type annotation. For supporting arguments we have a high degree of consensus, with Cohen's $\kappa$ ranges from 0.76 to 0.83 in all rounds and 0.80 overall. 
For argument type annotation, we achieve Cohen's $\kappa$ of 0.61 for \textsc{study}, 0.75 for \textsc{factual}, 0.71 for \textsc{opinion}, and 0.29 for \textsc{reasoning}\footnote{Many times annotators have different interpretation on \textsc{reasoning}, and frequently label it as \textsc{opinion}. This results in a low agreement for \textsc{reasoning}.}

\noindent \textbf{Statistics.} 
In total 995 sentences are identified as supporting arguments. 
Among those, 95 (9.55\%) are labeled as \textsc{study}, 497 (49.95\%) as \textsc{factual}, 363 (36.48\%) as \textsc{opinion}, and 40 (4.02\%) as \textsc{reasoning}.

We further analyze the source of the supporting arguments. Domain names of the citation articles are collected based on their URL, and then categorized into ``news", ``organization", ``scientific", ``blog", ``reference", and others, according to a taxonomy provided by Alexa\footnote{http://www.alexa.com/topsites/category} with a few edits to fit our dataset. 

\begin{figure}
    \hspace{-5mm}
    \includegraphics[width=88mm]{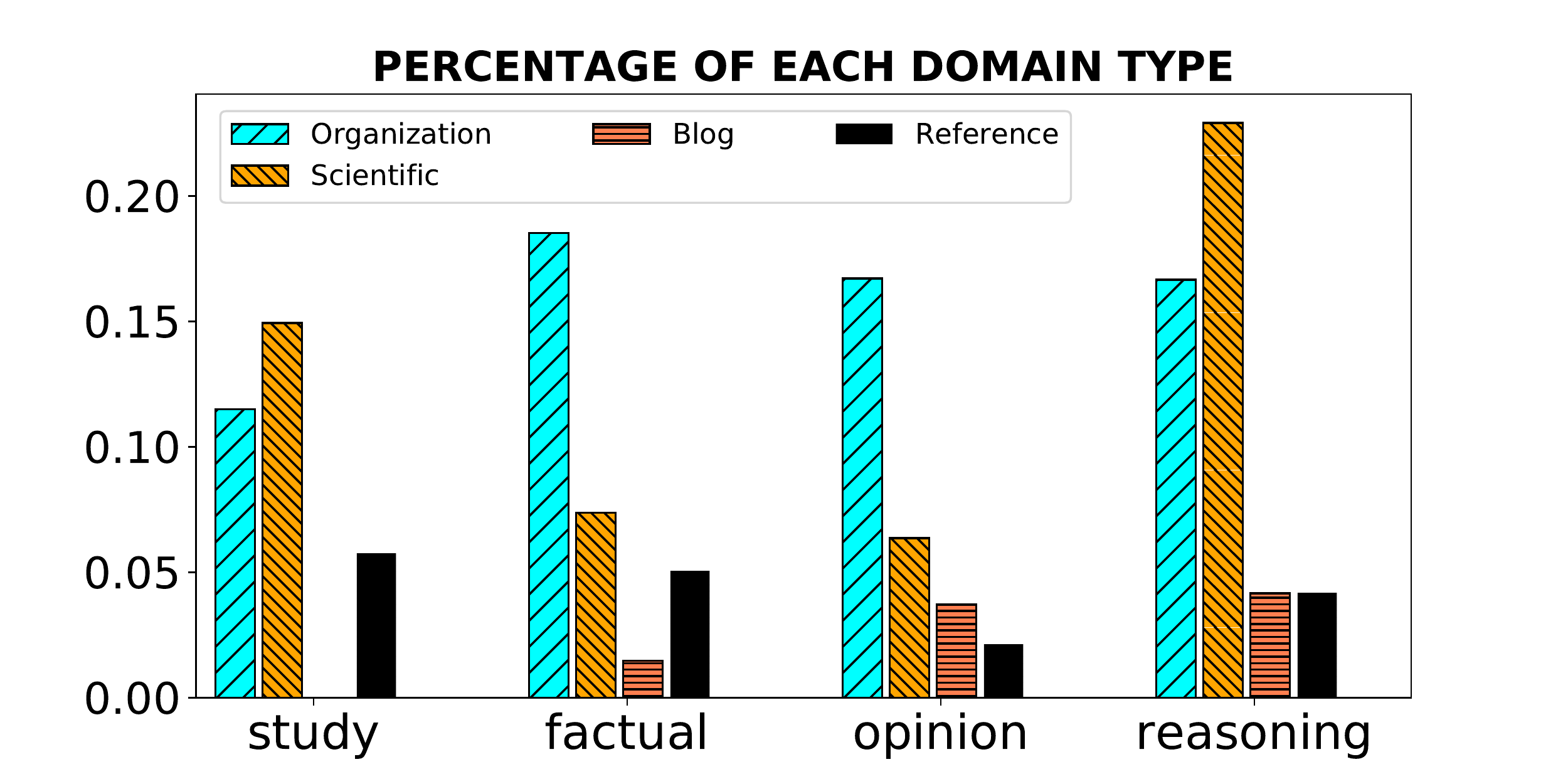}
    \vspace{-7mm}
    \caption{\fontsize{10}{12}\selectfont For each supporting argument type, from left to right shows the percentage of domain names of organizations, scientific, blog, and reference. We do not display statistics for news, because news articles take the same portion in all types (about 50\%).}
    \label{fig:type_domain_distribution}
\end{figure}

News articles are the major source for all types, which account for roughly 50\% for each. We show the distribution of other four types in Figure~\ref{fig:type_domain_distribution}. 
Arguments of \textsc{study} and \textsc{reasoning} are mostly from ``scientific" websites (14.9\% and 22.9\%), whereas ``organization" websites contribute a large portion of arguments of \textsc{factual} (18.5\%) and \textsc{opinion} (16.7\%).

\section{A Study On Argument Type Prediction}
\label{sec:type}
%Here we propose a feature-rich Log-linear classification model and a ranking model, together with naive baselines to study our hypothesis that type of sentence do affect its usage during debate construction. And also to provide some insights about what kind of information human debaters are looking for in the reference documents.

%Since type information is only labeled for supporting arguments but unavailable for all others. We need to automatically infer type in order to study the interaction between them.

%We propose a Log-linear classification model for this task. Since our type scheme is largely based on subjectivity and the source and form of information, features like named entity, concreteness, hedge cues and sentiment polarity would be helpful. We listed feature we used in table \ref{tab:features} in sentence level features. We use all labeled sentences and divide them into train (50\%), valid (25\%), test (25\%) sets for tuning and evaluation.

Here we characterize arguments of different types based on diverse features under the task of predicting argument types. Supporting arguments identified from previous section are utilized for experiments. We also leverage the learned classifier in this section to label the sentences that are not supporting arguments, which will be used for supporting argument detection in the next section. Four major types of features are considered.

%\vspace{1mm}
\noindent \textbf{Basic Features.} We calculate frequencies of unigram and bigram words, number of four major types of part-of-speech tags (verb, noun, adjective, and adverb), number of dependency relations, and number of seven types of named entities~\cite{chinchor1997muc}.

%\vspace{1mm}
\noindent \textbf{Sentiment Features.} We also compute number of positive, negative and neutral words in MPQA lexicon~\cite{wilson2005recognizing}, and number of words from a subset of semantic categories from General Inquirer~\cite{stone1966general}.\footnote{Categories used: Strong, Weak, Virtue, Vice, Ovrst (Overstated), Undrst (Understated), Academ (Academic), Doctrin (Doctrine), Econ\@ (Economic), Relig (Religious), Causal, Ought, and Perceiv (Perception).}

%\vspace{1mm}
\noindent \textbf{Discourse Features.} We use the number of discourse connectives from the top two levels of Penn Discourse Tree Bank~\cite{prasad2007penn}.

%\vspace{1mm}
\noindent \textbf{Style Features.} We measure word attributes for their concreteness (perceptible vs. conceptual), valence (or pleasantness), arousal (or intensity of emotion), and dominance (or degree of control) based on the lexicons collected by~\newcite{brysbaert2014concreteness} and~\newcite{warriner2013norms}.%\vspace{.1cm}

%\XY{The average concreteness score based on~\newcite{brysbaert2014concreteness}. And similarly Valence (Pleasantness, ``sunshine'' $>$ ``jail''), Arousal (Intensity of emotion, ``insanity'' $>$ ``calm''), Dominance (Degree of control, where ``completion'' $>$ ``dementia'') from \cite{warriner2013norms}}

\begin{figure}[t]
  \hspace{-4mm}
    \includegraphics[width=83mm]{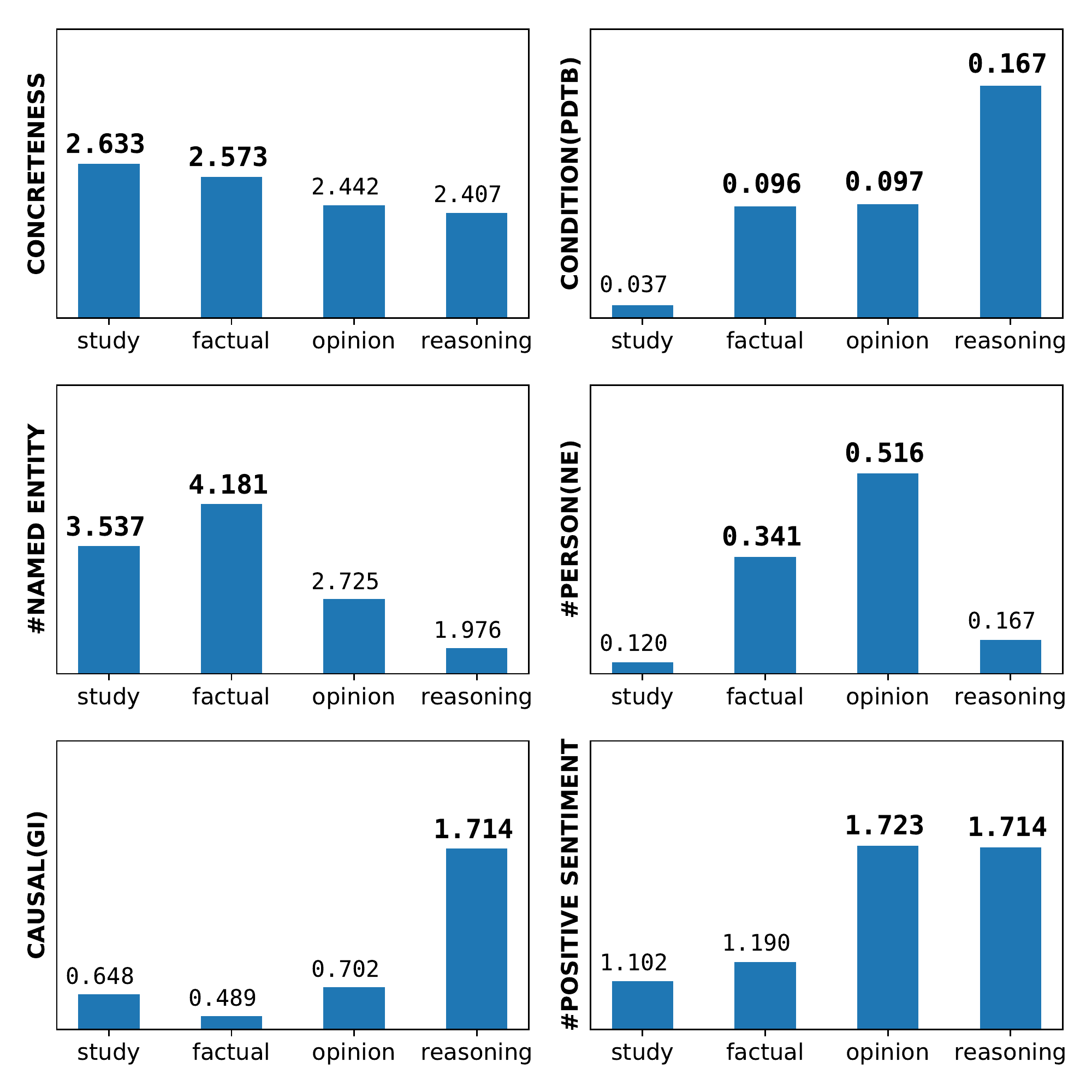}
  \vspace{-5mm}
  \caption{\fontsize{10}{12}\selectfont Average features values for different argument types. Numbers in \textbf{boldface} are significantly higher than the others based on paired $t$-test ($p<0.05$).}
%  Boldfaced numbers represent a ``group'' such that within the group there is no significant differences, throughout different groups there are significant differences based paired $t$-test ($p<0.05$).}
  \label{fig:pwttest}

\end{figure}

We utilize Log-linear model for argument type prediction with one-vs-rest setup. Three baselines are considered: (1) random guess, (2) majority class, and (3) unigrams and bigrams as features for Log-linear model. 
Identified supporting arguments are used for experiments, and divided into training set (50\%), validation set (25\%) and test set (25\%). From Table~\ref{tab:typeresult}, we can see that Log-linear model trained with all features outperforms the ones trained with ngram features. 
To further characterize arguments of different types, we display sample features with significant different values in Figure~\ref{fig:pwttest}. As can be seen, arguments of \textsc{study} and \textsc{factual} tend to contain more concrete words and named entities. Arguments of \textsc{opinion} mention more person names, which implies that expert opinions are commonly quoted.

\begin{table}[t]
    \centering
    \fontsize{10}{11}\selectfont
    \begin{tabular}{|l|c|c|}
    \hline
         & {\bf Acc} & {\bf F1}  \\
        \hline
        Majority class & 0.520 & 0.171 \\
        \hline
        Random & 0.240 & 0.199 \\
        \hline
        Log-linear (ngrams) & 0.535 & 0.277 \\
        \hline
        Log-linear (all features) & {\bf 0.622} & {\bf 0.436} \\
        \hline
    \end{tabular}
    \caption{\fontsize{10}{12}\selectfont Results for argument type prediction. One-vs-rest classifiers are learned for Log-linear models.}
    \label{tab:typeresult}
\end{table}

%Random guess has accuracy close to 1/4, which is intuitive since it is guessing one of the four types. Majority guess achieves much higher accuracy, which also makes sense since we notice that type factual is more prominent than other types in our dataset. N-gram model achieves slightly higher accuracy than majority guess, but much higher F1 score.
%Classification results are shown in Table \ref{tab:results}. A pilot study shows that  one-vs-rest outperforms multiclass classification. And by leveraging different features helps improve performance by large margin.
%

%Classification results show that Log-linear model trained with all features as one-vs-rest classifier has accuracy of 0.622, outperforming a Log-linear model trained with ngram features (unigrams and bigrams) with accuracy of 0.535. 
%To further characterize arguments of different types, we display sample features with significant different values in Figure~\ref{fig:pwttest}. 
%we conduct pair wise t-test on all features and listed major ones in figure \ref{fig:pwttest}. 
%
%As can be seen, arguments of \textsc{study} and \textsc{factual} tend to contain more concrete words and named entities. Arguments of \textsc{opinion} mention more person names, which implies that expert opinions are commonly quoted. 
%opinions usually precedes experts or authorities. %We suspect that human tend to synthesize results from multiple studies in one arguments, but only one opinion or reasoning at once. 
%And type reasoning has more words used in causal and conditional structures, which accords with our expectation.

%Now that we could detect the type of sentences, we may add this power to help infer which sentences could be used to construct arguments to support the claim. We name this process relevance inference, in the sense that our goldstandard data is annotated based on the relevance between sentences in the cited articles versus the sentences the author used with reference to that cited article. We first present a simple unsupervised baseline method, then discuss several ranking models and features we designed for them.

\section{Supporting Argument Detection}
\label{sec:relevance}
%The nature of this task can be abstract to ranking sentences by certain measurements, which potentially could provide us with some insights about how human beings select information during debates. Here we have xxx different types of features, and we further group them into xxx feature sets.

%The process of finding directly relevant sentence could be modeled as a ranking problem. In addition to features listed in section 4, we also consider similarity features and composite features as below:
%\XY{Due to the fact that most of the reference documents are closely related to the topic, and there are usually more than one relevant sentences but we only care about the most relevant one}
We cast the sentence-level supporting argument detection problem as a ranking task.\footnote{Many sentences in the citation article is relevant to the topic to various degrees. We focus on detecting the most relevant ones, and thus treat it as a ranking problem instead of a binary classification task.}  
Features in Section~\ref{sec:type} are also utilized here as ``Sentence features" with additional features considering the sentence position in the article. We further employ features that measure similarity between claims and sentences, and the composite features that leverage argument type information. 

\vspace{1mm}
\noindent \textbf{Similarity Features.} We compute similarity between claim and candidate sentence based on TF-IDF and average word embeddings. We also consider ROUGE~\cite{lin2004rouge}, a recall oriented metric for summarization evaluation. In particular, ROUGE-L, a variation based on longest common subsequence, is computed by treating claim as reference and each candidate sentence as sample summary. In similar manner we use BLEU~\cite{papineni2002bleu}, a precision oriented metric.

\vspace{1mm}
\noindent \textbf{Composite Features.} We adopt composite features to study the interaction of other features with type of the sentence. Given claim $c$ and sentence $s$ with any feature mentioned above, a composite feature function $\phi_{M\texttt{\textbf{\scriptsize (type, feature)}}}(s,c)$ is set to the actual feature value if and only if the argument type matches. For instance, if the ROUGE-L score is 0.2, and $s$ is of type \textsc{study}, then {\small $\phi_{M\texttt{\textbf{\scriptsize (study, ROUGE)}}}(s,c)=0.2$ }\\
{\small $\phi_{M\texttt{\textbf{\scriptsize (factual, ROUGE)}}}{\small(s,c)}$, $\phi_{M\texttt{\textbf{\scriptsize (opinion, ROUGE)}}}{\small(s,c)}$, $\phi_{M\texttt{\textbf{\scriptsize (reasoning, ROUGE)}}}(s,c)$} are all set to 0.

%features in Table \ref{tab:features} and try different combination of them when composed with type information. Table \ref{tab:featset} shows the basic feature sets we used. Here \textbf{Comp} stands for composite feature. For example, if we compose sentence level features with type, suppose the dimension for sentence level features is $\mathcal{D}$, then composite feature has dimension $4\mathcal{D}$. Only the dimensions corresponding to one of the four predicted type will be actual values for features, other dimensions will be set to zero. By doing this we could learn the interaction of features and types.

We choose LambdaMART~\cite{burges2010ranknet} for experiments, which is shown to be successful for many text ranking problems~\cite{chapelle2011yahoo}. Our model is evaluated by Mean Reciprocal Rank (MRR) and Normalized Discounted Cumulative Gain (NDCG) using 5-fold cross validation. We compare to TFIDF and Word embedding similarity baselines, and LambdaMART trained with ngrams (unigrams and bigrams). 

%\vspace{-3mm}
\begin{table}[t]
{
%\hspace{4mm}
\fontsize{9}{10}\selectfont
\setlength{\tabcolsep}{0.6mm}
\begin{tabular}{|l|c|c|}
    \hline
        Feature set & \textbf{MRR} & \textbf{NDCG} \\ \hline
        \multicolumn{1}{|l|}{\textbf{Baselines}} & &\\ 
        TFIDF similarity & 45.48 & 56.48 \\ 
        W2V similarity & 47.65 & 59.00 \\ 
        Ngrams & 27.26 & 43.83 \\ \hline \hline
        
        \multicolumn{1}{|l|}{\textbf{Separate feature sets}} & & \\ 
        Sentence (Sen) & 55.38* & 65.09* \\ 
        Similarity (Simi) & 43.13 & 55.16 \\ 
        Comp(type, Sen) + Comp(type, Simi) & 55.75* & 64.91* \\ \hline 
%        Comp(type,Sen) + Comp(type,Sim) + Comp(type,Claim) & 55.80* & 64.85*  \\ \hline
        
        \multicolumn{1}{|l|}{\textbf{Additive Feature Test}} & & \\ 
        Sen + Ngrams + Simi  & 56.43* & 65.79* \\ 
        ~~~~ + Comp(type, Sen) + Comp(type, Simi) & \textbf{57.65*} & \textbf{66.51*} \\ 
        ~~~~ + Comp(type, Claim) & 56.58* & 65.68* \\ \hline
\end{tabular}
%\vspace{-3mm}
\caption{\fontsize{10}{12}\selectfont Supporting argument detection results. %``Sentence" stands for features discussed in section 4 for modeling candidate sentences. ``Similarity" refers to similarity features. 
Comp(type, Sen) stands for composite features of argument type and sentence features, similarly for Comp(type,Simi).  Comp(type,Claim) represents composite features of type and claim features. Results that are statistically significantly better than all three baselines are marked with $\ast$ (paired $t$-test, $p<0.05$).}
\label{tab:mainresult}
}
\end{table}

Results in Table~\ref{tab:mainresult} show that %Comp(type, Sen) stands for composite features of argument type and sentence features, similarly for Comp(type, Simi). We also compute sentence features for the given claim, and use Comp(type, Claim) to characterize the interplay between claim and argument types. 
using composite features with argument type information (Comp(type, Sen) + Comp(type, Simi)) can improve the ranking performance. Specifically, the best performance is achieved by adding composite features to sentence features, similarity features, and ngram features.
As can be seen, supervised methods outperform unsupervised baseline methods. And similarity features have similar performance as those baselines. The best performance is achieved by combination of sentence features, N-grams, similarity, and two composite types, which is boldfaced. Feature sets that significantly outperform all three baselines are marked with $\ast$.

%To show the difficulties of detecting each individual types, we breakdown ranking results by types for the best performing system in \ref{tab:mainresult}.
%From \ref{tab:breakdown} we can see, type study is easiest to detect and type reasoning is the hardest one, which is consistent with Inter-annotator agreement results on different types.

For feature analysis, we conduct $t$-test for individual feature values between supporting arguments and the others. We breakdown features according to their argument types and show top salient composite features in Table \ref{tab:significance}. 
For all sentences of type \textsc{study}, relevant ones tend to contain more ``percentage" and more concrete words. We also notice those sentences with more hedging words are more likely to be considered. For sentences of \textsc{factual}, position of sentence in article plays an important role, as well as their similarity to the claim based on ROUGE scores. For type \textsc{opinion}, unlike all other types, position of sentence seems to be insignificant. As we could imagine, opinionated information might scatter around the whole documents. For sentences of \textsc{reasoning}, the ones that can be used as supporting arguments tend to be less concrete and less emotional, as opposed to opinion. 

\begin{table}[t]
{
\fontsize{8.5}{9}\selectfont
\setlength{\tabcolsep}{1.1mm}

\begin{tabular}{|p{1.5cm}|p{1.1cm}|p{1.1cm}|p{1.1cm}|p{1.3cm}|}
    \hline
    \multicolumn{1}{|c|}{Feature} & 
    \multicolumn{1}{c|}{\textsc{Study}} & 
    \multicolumn{1}{c|}{\textsc{Factual}} & 
    \multicolumn{1}{c|}{\textsc{Opinion}} & 
    \multicolumn{1}{c|}{\textsc{Reasoning}}  \\ \hline
    
    %\# PERCENT, NE & $\ast\ast\uparrow\uparrow\uparrow\uparrow$ & $\circ$ & $\circ$ & $\ast\uparrow\uparrow$  \\ \hline
    
    \# PERC, NE & $\ast\ast\uparrow\uparrow\uparrow\uparrow$ & -- & -- & --  \\ \hline
    
    %\# LOCATION, NE &  $\ast\uparrow\uparrow\uparrow\uparrow$& $\ast\ast\uparrow\uparrow$ & $\circ$ & $\ast\ast\uparrow$  \\ \hline
    
    \# LOC, NE & -- & $\ast\ast\uparrow\uparrow$ & -- & $\ast\ast\uparrow$  \\ \hline
    
    position \newline of sentence & $\ast\ast\downarrow\downarrow$ &  $\ast\ast\ast\ast$ $\downarrow\downarrow$ & -- & $\ast\ast\ast\ast$ $\downarrow\downarrow\downarrow\downarrow$ \\ \hline
    
    %concreteness \newline of sentence&  $\ast\ast\ast\uparrow\uparrow$ & $\ast\uparrow$ & $\ast\ast\uparrow\uparrow$ & $\ast\ast\ast\downarrow$  \\ \hline
    concreteness \newline of sentence&  $\ast\ast\ast$ $\uparrow\uparrow$ & -- & $\ast\ast\uparrow\uparrow$ & $\ast\ast\ast\downarrow$  \\ \hline
    
  %  \# pragmatic condition(PDTB) & -- & -- & -- & 
    
    %arousal of sentence & $\ast\ast\ast\uparrow\uparrow$ & $\ast\uparrow$ & $\ast\ast\uparrow\uparrow$ & $\ast\ast\downarrow$  \\ \hline
    
    arousal \newline of sentence & $\ast\ast\ast$ $\uparrow\uparrow$ & -- & $\ast\ast\uparrow\uparrow$ & $\ast\ast\downarrow$  \\ \hline
    
    \# hedging \newline word & $\ast\ast\uparrow\uparrow\uparrow$ & -- & -- & --  \\ \hline
    
    ROUGE & $\ast\ast\ast$$\uparrow\uparrow$ & $\ast\ast\ast\uparrow$ & $\ast\ast\uparrow\uparrow$ & --  \\ \hline
    
    %\# neutral word \newline (MPQA) & $\ast\uparrow\uparrow$& $\circ$ & $\ast\ast\uparrow\uparrow$ & $\ast\ast\ast\downarrow$\\\hline
   % \# neutral word \newline (MPQA) & -- & -- & -- & $\ast\ast\ast\downarrow$\\\hline 
    
    concreteness \newline of claim&  $\ast\ast\ast$ $\uparrow\uparrow$ & -- & $\ast\ast\uparrow\uparrow$ & $\ast\ast\ast\downarrow$  \\ \hline
    arousal \newline of claim & $\ast\ast\ast$ $\uparrow\uparrow$ & -- & $\ast\ast\uparrow\uparrow$ & $\ast\ast\ast\downarrow$  \\ \hline
    
\end{tabular}
%\vspace{-3mm}
\caption{\fontsize{10}{12}\selectfont Comparison of feature significance under composition with different types. The number of $\ast$ stands for the p-value based on $t$-test between supporting argument sentences and the others after Bonferroni correction. From one $\ast$ to four, the $p$-value scales as: 0.05, 1e-3, 1e-5, and 1e-10. When mean value of supporting argument sentences is larger, $\uparrow$ is used; otherwise, $\downarrow$ is displayed. Number of arrows represents the ratio of the larger value over smaller one. ``-" indicates no significant difference.} 
\label{tab:significance}
\vspace{-.5cm}
}
\end{table}

\section{Conclusion}
\label{sec:conclusion}
We presented a novel study on the task of sentence-level supporting argument detection from relevant documents for a user-specified claim. Based on our newly-collected dataset, we characterized arguments of different types with a rich feature set. We also showed that leveraging argument type information can further improve the performance of supporting argument detection.

%In this paper we study how to identify supporting arguments under debate settings by leveraging argument type information. And we annotate a debate corpus with sentence level type information to facilitate automatic detection. Experimental results indicate the type information can help improve detection performance. 

%Due to the difficulties lie in the nature of this problem, there are still many interesting aspects worth pursing in the future. For example, how human edits the original sentences into different variations; how to better utilize the context information to identify the source, and the techniques used to synthesize different arguments into a coherent paragraph.

%the techniques people use to synthesize different arguments into a coherent paragraph; the importance of context information surrounding the directly relevant sentence; and how to automatically identify resources as potential citation documents.
%%% future
% add context information for detecting relevant sentences

% study how human uses the sentences from citation articles

\section*{Acknowledgments}
This work was supported in part by National Science Foundation Grant IIS-1566382 and a GPU gift from Nvidia. We thank Kechen Qin for his help on data collection. We also appreciate the valuable suggestions on various aspects of this work from three anonymous reviewers.

\newpage

\bibliography{argument,additional}

\begin{thebibliography}{}
\expandafter\ifx\csname natexlab\endcsname\relax\def\natexlab#1{#1}\fi

\bibitem[{Al~Khatib et~al.(2016)Al~Khatib, Wachsmuth, Kiesel, Hagen, and
  Stein}]{alkhatib-EtAl:2016:COLING}
Khalid Al~Khatib, Henning Wachsmuth, Johannes Kiesel, Matthias Hagen, and Benno
  Stein. 2016.
\newblock A news editorial corpus for mining argumentation strategies.
\newblock In {\em Proceedings of COLING 2016, the 26th International Conference
  on Computational Linguistics: Technical Papers\/}. Osaka, Japan, pages
  3433--3443.

\bibitem[{Biran and Rambow(2011)}]{biran2011identifying}
Or~Biran and Owen Rambow. 2011.
\newblock Identifying justifications in written dialogs by classifying text as
  argumentative.
\newblock {\em International Journal of Semantic Computing\/} 5(04):363--381.

\bibitem[{Brysbaert et~al.(2014)Brysbaert, Warriner, and
  Kuperman}]{brysbaert2014concreteness}
Marc Brysbaert, Amy~Beth Warriner, and Victor Kuperman. 2014.
\newblock Concreteness ratings for 40 thousand generally known english word
  lemmas.
\newblock {\em Behavior research methods\/} 46(3):904--911.

\bibitem[{Burges(2010)}]{burges2010ranknet}
Christopher~JC Burges. 2010.
\newblock From ranknet to lambdarank to lambdamart: An overview.
\newblock {\em Learning\/} 11(23-581):81.

\bibitem[{Chapelle and Chang(2011)}]{chapelle2011yahoo}
Olivier Chapelle and Yi~Chang. 2011.
\newblock Yahoo! learning to rank challenge overview.
\newblock In {\em Yahoo! Learning to Rank Challenge\/}. pages 1--24.

\bibitem[{Chinchor and Robinson(1997)}]{chinchor1997muc}
Nancy Chinchor and Patricia Robinson. 1997.
\newblock Muc-7 named entity task definition.
\newblock In {\em Proceedings of the 7th Conference on Message
  Understanding\/}. volume~29.

\bibitem[{Feng and Hirst(2011)}]{feng2011classifying}
Vanessa~Wei Feng and Graeme Hirst. 2011.
\newblock Classifying arguments by scheme.
\newblock In {\em Proceedings of the 49th Annual Meeting of the Association for
  Computational Linguistics: Human Language Technologies-Volume 1\/}.
  Association for Computational Linguistics, pages 987--996.

\bibitem[{Habernal and Gurevych(2015)}]{habernal-gurevych:2015:EMNLP}
Ivan Habernal and Iryna Gurevych. 2015.
\newblock Exploiting debate portals for semi-supervised argumentation mining in
  user-generated web discourse.
\newblock In {\em Proceedings of the 2015 Conference on Empirical Methods in
  Natural Language Processing\/}. Association for Computational Linguistics,
  Lisbon, Portugal, pages 2127--2137.

\bibitem[{Lin(2004)}]{lin2004rouge}
Chin-Yew Lin. 2004.
\newblock Rouge: A package for automatic evaluation of summaries.
\newblock In {\em Text summarization branches out: Proceedings of the ACL-04
  workshop\/}. Barcelona, Spain, volume~8.

\bibitem[{Mochales and Moens(2011)}]{mochales2011argumentation}
Raquel Mochales and Marie-Francine Moens. 2011.
\newblock Argumentation mining.
\newblock {\em Artificial Intelligence and Law\/} 19(1):1--22.

\bibitem[{Moens et~al.(2007)Moens, Boiy, Palau, and Reed}]{moens2007automatic}
Marie-Francine Moens, Erik Boiy, Raquel~Mochales Palau, and Chris Reed. 2007.
\newblock Automatic detection of arguments in legal texts.
\newblock In {\em Proceedings of the 11th international conference on
  Artificial intelligence and law\/}. ACM, pages 225--230.

\bibitem[{Nguyen and Litman(2016)}]{nguyen-litman:2016:P16-1}
Huy Nguyen and Diane Litman. 2016.
\newblock Context-aware argumentative relation mining.
\newblock In {\em Proceedings of the 54th Annual Meeting of the Association for
  Computational Linguistics (Volume 1: Long Papers)\/}. Association for
  Computational Linguistics, Berlin, Germany, pages 1127--1137.

\bibitem[{Palau and Moens(2009)}]{palau2009argumentation}
Raquel~Mochales Palau and Marie-Francine Moens. 2009.
\newblock Argumentation mining: the detection, classification and structure of
  arguments in text.
\newblock In {\em Proceedings of the 12th international conference on
  artificial intelligence and law\/}. ACM, pages 98--107.

\bibitem[{Papineni et~al.(2002)Papineni, Roukos, Ward, and
  Zhu}]{papineni2002bleu}
Kishore Papineni, Salim Roukos, Todd Ward, and Wei-Jing Zhu. 2002.
\newblock Bleu: a method for automatic evaluation of machine translation.
\newblock In {\em Proceedings of the 40th annual meeting on association for
  computational linguistics\/}. Association for Computational Linguistics,
  pages 311--318.

\bibitem[{Park and Cardie(2014)}]{park2014identifying}
Joonsuk Park and Claire Cardie. 2014.
\newblock Identifying appropriate support for propositions in online user
  comments.
\newblock In {\em Proceedings of the First Workshop on Argumentation Mining\/}.
  pages 29--38.

\bibitem[{Prasad et~al.(2007)Prasad, Miltsakaki, Dinesh, Lee, Joshi, Robaldo,
  and Webber}]{prasad2007penn}
Rashmi Prasad, Eleni Miltsakaki, Nikhil Dinesh, Alan Lee, Aravind Joshi, Livio
  Robaldo, and Bonnie~L Webber. 2007.
\newblock The penn discourse treebank 2.0 annotation manual .

\bibitem[{Rieke et~al.(1997)Rieke, Sillars, and
  Peterson}]{rieke1997argumentation}
Richard~D Rieke, Malcolm~Osgood Sillars, and Tarla~Rai Peterson. 1997.
\newblock {\em Argumentation and critical decision making\/}.
\newblock New York: Longman.

\bibitem[{Rinott et~al.(2015)Rinott, Dankin, Perez, Khapra, Aharoni, and
  Slonim}]{rinott2015show}
Ruty Rinott, Lena Dankin, Carlos~Alzate Perez, Mitesh~M Khapra, Ehud Aharoni,
  and Noam Slonim. 2015.
\newblock Show me your evidence-an automatic method for context dependent
  evidence detection.
\newblock In {\em EMNLP\/}. pages 440--450.

\bibitem[{Rooney et~al.(2012)Rooney, Wang, and Browne}]{rooney2012applying}
Niall Rooney, Hui Wang, and Fiona Browne. 2012.
\newblock Applying kernel methods to argumentation mining.
\newblock In {\em Twenty-Fifth International FLAIRS Conference\/}.

\bibitem[{Stab and Gurevych(2014)}]{stab2014identifying}
Christian Stab and Iryna Gurevych. 2014.
\newblock Identifying argumentative discourse structures in persuasive essays.
\newblock In {\em EMNLP\/}. pages 46--56.

\bibitem[{Stone et~al.(1966)Stone, Dunphy, and Smith}]{stone1966general}
Philip~J Stone, Dexter~C Dunphy, and Marshall~S Smith. 1966.
\newblock The general inquirer: A computer approach to content analysis. .

\bibitem[{Warriner et~al.(2013)Warriner, Kuperman, and
  Brysbaert}]{warriner2013norms}
Amy~Beth Warriner, Victor Kuperman, and Marc Brysbaert. 2013.
\newblock Norms of valence, arousal, and dominance for 13,915 english lemmas.
\newblock {\em Behavior research methods\/} 45(4):1191--1207.

\bibitem[{Wilson et~al.(2005)Wilson, Wiebe, and
  Hoffmann}]{wilson2005recognizing}
Theresa Wilson, Janyce Wiebe, and Paul Hoffmann. 2005.
\newblock Recognizing contextual polarity in phrase-level sentiment analysis.
\newblock In {\em Proceedings of the conference on human language technology
  and empirical methods in natural language processing\/}. Association for
  Computational Linguistics, pages 347--354.

\end{thebibliography}
\bibliographystyle{acl_natbib}

\end{document}